\documentclass[10pt,twocolumn,letterpaper]{article}

\usepackage{btas}
\usepackage{times}
\usepackage{epsfig}
\usepackage{graphicx}
\usepackage{amsmath}
\usepackage{amssymb}
\usepackage{booktabs}
\usepackage{bbding}
\usepackage{subfigure}
\usepackage{diagbox}
\usepackage{textcomp}
\usepackage{setspace}
\usepackage[pagebackref=true,breaklinks=true,letterpaper=true,colorlinks,bookmarks=false]{hyperref}

\btasfinalcopy 

\ifbtasfinal\fi
\begin{document}
\hyphenpenalty=400

\title{FaceBoxes: A CPU Real-time Face Detector with High Accuracy}

\author{Shifeng Zhang\quad Xiangyu Zhu\quad Zhen Lei\quad Hailin Shi\quad Xiaobo Wang\quad Stan Z. Li\\
CBSR \& NLPR, Institute of Automation, Chinese Academy of Sciences, Beijing, China\\
University of Chinese Academy of Sciences, Beijing, China\\
{\tt\small \{shifeng.zhang,xiangyu.zhu,zlei,hailin.shi,xiaobo.wang,szli\}@nlpr.ia.ac.cn}
}
\maketitle
\thispagestyle{empty}

\begin{abstract}
Although tremendous strides have been made in face detection, one of the remaining open challenges is to achieve real-time speed on the CPU as well as maintain high performance, since effective models for face detection tend to be computationally prohibitive. To address this challenge, we propose a novel face detector, named FaceBoxes, with superior performance on both speed and accuracy. Specifically, our method has a lightweight yet powerful network structure that consists of the Rapidly Digested Convolutional Layers (RDCL) and the Multiple Scale Convolutional Layers (MSCL). The RDCL is designed to enable FaceBoxes to achieve real-time speed on the CPU. The MSCL aims at enriching the receptive fields and discretizing anchors over different layers to handle faces of various scales. Besides, we propose a new anchor densification strategy to make different types of anchors have the same density on the image, which significantly improves the recall rate of small faces. As a consequence, the proposed detector runs at $20$ FPS on a single CPU core and $125$ FPS using a GPU for VGA-resolution images. Moreover, the speed of FaceBoxes is invariant to the number of faces. We comprehensively evaluate this method and present state-of-the-art detection performance on several face detection benchmark datasets, including the AFW, PASCAL face, and FDDB. Code is available at \url{https://github.com/sfzhang15/FaceBoxes}. 
\end{abstract}

\setlength{\parskip}{-0.6\baselineskip}
\section{Introduction}
\setlength{\parskip}{-0.3\baselineskip}
Face detection is one of the fundamental problems in computer vision and pattern recognition. It plays an important role in many subsequent face-related applications, such as face alignment~\cite{zhu2016face}, face recognition~\cite{zhu2015high} and face tracking~\cite{kim2008face}. With the great progress over the past few decades, especially the breakthrough of convolutional neural network, face detection has been successfully applied in our daily life under various scenarios.
\setlength{\parskip}{-0.0\baselineskip}

However, there are still some tough challenges in uncontrolled face detection problem, especially for the CPU devices. The challenges mainly come from two requirements for face detectors: 1) The large visual variation of faces in the cluttered backgrounds requires face detectors to accurately address a complicated face and non-face classification problem; 2) The large search space of possible face positions and face sizes further imposes a time efficiency requirement. These two requirements are conflicting, since high-accuracy face detectors tend to be computationally expensive. Therefore, it is one of the remaining open issues for practical face detectors on the CPU devices to achieve real-time speed as well as maintain high performance.

In order to meet these two conflicting requirements, face detection has been intensely studied mainly in two ways. The early way is based on hand-craft features. Following the pioneering work of Viola-Jones face detector~\cite{viola2004robust}, most of the early works focus on designing robust features and training effective classifiers. Besides the cascade structure, the deformable part model (DPM) is introduced into face detection tasks and achieves remarkable performance. However, these methods highly depend on non-robust hand-craft features and optimize each component separately, making the face detection pipeline sub-optimal. In brief, they are efficient on the CPU but not accurate enough against the large visual variation of faces.

The other way is based on the convolutional neural network (CNN) which has achieved remarkable successes in recent years, ranging from image classification to object detection. Recently, CNN has been successfully introduced into the face detection task as feature extractor in the traditional face detection framewrok~\cite{ohn2017boost,yang2015convolutional,yang2015facial}. Moreover, some face detectors~\cite{chen2016supervised,zhu2016cms} have inherited valid techniques from the generic object detection methods, such as Faster R-CNN~\cite{ren2015faster}. These CNN based face detection methods are robust to the large variation of facial appearances and demonstrate state-of-the-art performance. But they are too time-consuming to achieve real-time speed, especially on the CPU devices.

These two ways have their own advantages. The former has fast speed while the latter owns high accuracy. To perform well on both speed and accuracy, one natural idea is to combine the advantages of these two types of methods. Therefore, cascaded CNN based methods~\cite{li2015convolutional,zhang2016joint} are proposed to put features learned by CNN into cascade framework in order to boost the performance and keep efficient. However, there are three problems in cascaded CNN based methods: 1) Their speed is negatively related to the number of faces on the image. The speed would dramatically degrade as the number of faces increases; 2) The cascade based detectors optimize each component separately, making the training process extremely complicated and the final model sub-optimal; 3) For the VGA-resolution images, their runtime efficiency on the CPU is about 14 FPS, which is not fast enough to reach the real-time speed.

In this paper, inspired by the RPN in Faster R-CNN~\cite{ren2015faster} and the multi-scale mechanism in SSD~\cite{liu2016ssd}, we develop a state-of-the-art face detector with real-time speed on the CPU. Specifically, we propose a novel face detector named FaceBoxes, which only contains a single fully convolutional neural network and can be trained end-to-end. The proposed method has a lightweight yet powerful network structure (as shown in Fig.~\ref{fig:FaceBoxes}) that consists of the Rapidly Digested Convolutional Layers (RDCL) and the Multiple Scale Convolutional Layers (MSCL). The RDCL is designed to enable FaceBoxes to achieve real-time speed on the CPU, and the MSCL aims at enriching the receptive fields and discretizing anchors over different layers to handle various scales of faces. Besides, we propose a new anchor densification strategy to make different types of anchors have the same density on the input image, which significantly improves the recall rate of small faces. Consequently, for VGA-resolution images, our face detector runs at $20$ FPS on a single CPU core and $125$ FPS using a GPU. More importantly, the speed of FaceBoxes is invariant to the number of faces on the image. We comprehensively evaluate this method and demonstrate state-of-the-art detection performance on several face detection benchmark datasets, including the AFW, PASCAL face, and FDDB. 

For clarity, the main contributions of this work can be summarized as four-fold:
\setlength{\parskip}{-0.75\baselineskip}
\begin{itemize}
\setlength{\itemsep}{0pt}
\setlength{\parsep}{0pt}
\setlength{\parskip}{1pt}
\item We design the Rapidly Digested Convolutional Layers (RDCL) to enable face detection to achieve real-time speed on the CPU;
\item We introduce the Multiple Scale Convolutional Layers (MSCL) to handle various scales of face via enriching receptive fields and discretizing anchors over layers.
\item We present a new anchor densification strategy to improve the recall rate of small faces;
\item We further improve the state-of-the-art performance on the AFW, PASCAL face, and FDDB datasets.
\end{itemize}
\setlength{\parskip}{-0.5\baselineskip}

The rest of the paper is organized as follows. Section~\ref{2} reviews the related work. Analysis of the FaceBoxes is presented in section~\ref{3}. Section~\ref{4} shows the experimental results and section~\ref{5} concludes the paper.
\setlength{\parskip}{0\baselineskip}
\section{Related work} \label{2}
Modern face detection approaches can be roughly divided into two different categories. One is based on hand-craft features, and the other one is built on CNN. This section briefly reviews these two kinds of methods.

\subsection{Hand-craft based methods}
Previous face detection systems are mostly based on hand-craft features. Since the seminal Viola-Jones face detector~\cite{viola2004robust} that proposes to combine Haar feature, Adaboost learning and cascade inference for face detection, many subsequent works are proposed for real-time face detection, such as new local features~\cite{liao2016fast,yang2014aggregate}, new boosting algorithms~\cite{brubaker2008design,pham2007fast} and new cascade structures~\cite{bourdev2005robust,li2002statistical}.

Besides the cascade framework, methods based on structural models progressively achieve better performance and become more and more efficient. Some researches~\cite{yan2014fastest,yan2014face,zhu2012face} introduce the deformable part model (DPM) into face detection tasks. These works use supervised parts, more pose partition, better training or more efficient inference to achieve remarkable detection performance.

\subsection{CNN based methods}
The first use of CNN for face detection can be traced back to 1994. Vaillant et al.~\cite{vaillant1994original} use a trained CNN in a sliding windows manner to detect faces. Rowley et al.~\cite{rowley1998neural,rowley1998rotation} introduce a retinally connected neural network for upright frontal face detection, and a ``router" network designed to estimate the orientation for rotation invariant face detection. Garcia et al.~\cite{garcia2002neural} develop a neural network to detect semi-frontal faces. Osadchy et al.~\cite{osadchy2007synergistic} train a CNN for simultaneous face detection and pose estimation. These earlier methods can get relatively good performance only on easy dataset. 

Recent years have witnessed the advance of CNN based face detectors. CCF~\cite{yang2015convolutional} uses boosting on top of CNN features for face detection. Farfade et al.~\cite{farfade2015multi} fine-tune CNN model trained on 1k ImageNet classification task for face and non-face classification task. Faceness~\cite{yang2015facial} trains a series of CNNs for facial attribute recognition to detect partially occluded faces. CascadeCNN~\cite{li2015convolutional} develops a cascade architecture built on CNNs with powerful discriminative capability and high performance. Qin et al.~\cite{qin2016joint} propose to jointly train CascadeCNN to realize end-to-end optimization. Similar to~\cite{chen2014joint}, MTCNN~\cite{zhang2016joint} proposes a multi-task cascaded CNNs based framework for joint face detection and alignment. UnitBox~\cite{yu2016unitbox} introduces a new intersection-over-union loss function. CMS-RCNN~\cite{zhu2016cms} uses Faster R-CNN in face detection with body contextual information. Convnet~\cite{li2016face} integrates CNN with 3D face model in an end-to-end multi-task learning framework. STN~\cite{chen2016supervised} proposes a new supervised transformer network and a ROI convolution for face detection. 

\begin{figure*}[htbp]
\centering
\includegraphics[width=1.0\textwidth]{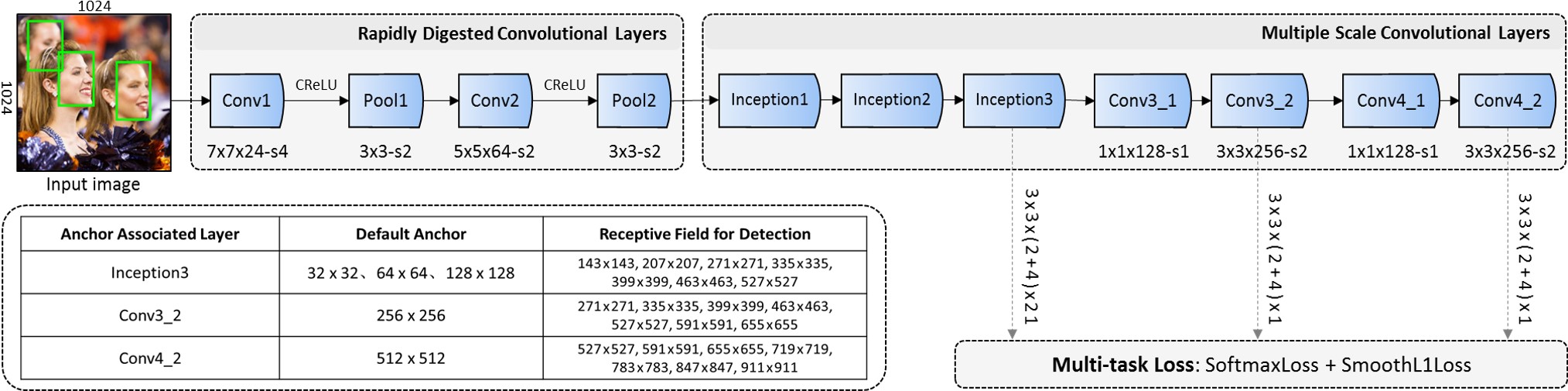}
\caption{Architecture of the FaceBoxes and the detailed information table about our anchor designs.} \label{fig:FaceBoxes}
\end{figure*}

\section{FaceBoxes} \label{3}
\setlength{\parskip}{-0.4\baselineskip}
This section presents our three contributions that make the FaceBoxes accurate and efficient on the CPU devices: the Rapidly Digested Convolutional Layers (RDCL), the Multiple Scale Convolutional Layers (MSCL) and the anchor densification strategy. Finally, we introduce the associated training methodology.
\setlength{\parskip}{-0.4\baselineskip}

\subsection{Rapidly Digested Convolutional Layers}
\setlength{\parskip}{-0.4\baselineskip}
Most of the CNN based face detection methods are usually limited by the heavy cost of time, especially on the CPU devices. More precisely, the convolution operation for CPU is extremely time-consuming when the size of input, kernel and output are large. Our RDCL is designed to fast shrink the input spatial size by suitable kernel size with reducing the number of output channels, enabling the FaceBoxes to reach real-time speed on the CPU devices as follows:
\setlength{\parskip}{-0.6\baselineskip}
\begin{itemize}
\setlength{\itemsep}{0pt}
\setlength{\parsep}{0pt}
\setlength{\parskip}{0pt}
\item \textbf{Shrinking the spatial size of input}: To rapidly shrink the spatial size of input, our RDCL sets a series of large stride sizes for its convolution and pooling layers. As illustrated in Fig.~\ref{fig:FaceBoxes}, the stride size of Conv1, Pool1, Conv2 and Pool2 are $4$, $2$, $2$ and $2$, respectively. The total stride size of RDCL is $32$, which means the input spatial size is reduced by $32$ times quickly.
\item \textbf{Choosing suitable kernel size}: The kernel size of the first few layers in one network should be small so as to speed up, while it is also supposed to be large enough to alleviate the information loss brought by the spatial size reducing. As shown in Fig.~\ref{fig:FaceBoxes}, to keep efficient as well as effective, we choose $7\times7$, $5\times5$ and $3\times3$ kernel size for Conv1, Conv2 and all Pool layers, respectively.
\item \textbf{Reducing the number of output channels}: We utilize the C.ReLU activation function (illustrated in Fig.~\ref{fig:CReLU}) to reduce the number of output channels. C.ReLU~\cite{shang2016understanding} is motivated from the observation in CNN that the filters in the lower layers form pairs (\textit{i.e.}, filters with opposite phase). From this observation, C.ReLU can double the number of output channels by simply concatenating negated outputs before applying ReLU. Using C.ReLU significantly increases speed with negligible decline in accuracy.
\end{itemize}
\setlength{\parskip}{0\baselineskip}

\subsection{Multiple Scale Convolutional Layers}
\begin{figure}[htbp]
\centering
\subfigure[]{
\label{fig:CReLU}
\includegraphics[width=3.05cm]{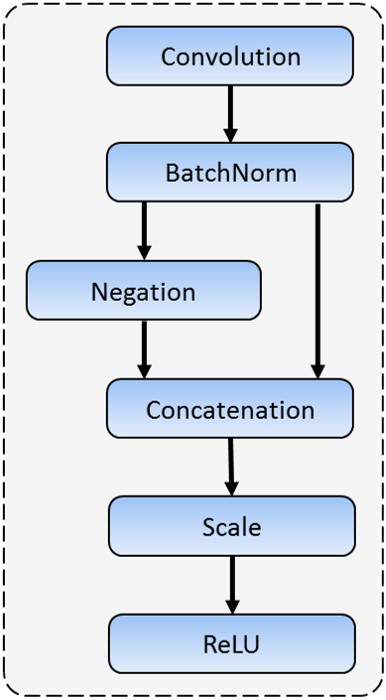}}
\subfigure[]{
\label{fig:Inception}
\includegraphics[width=4.95cm]{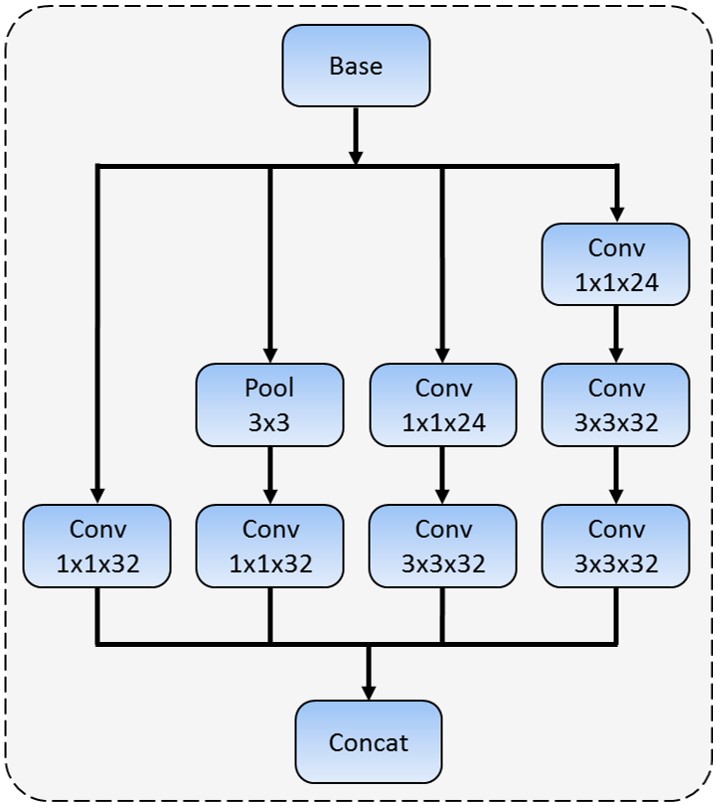}}
\caption{(a) The C.ReLU modules where Negation simply multiplies $-1$ to the output of Convolution. (b) The Inception modules.}
\end{figure}

The proposed method is based on RPN which is developed as a class-agnostic proposer in the scenario of multi-category object detection. For the single-category detection task (\textit{e.g.}, face detection), RPN is naturally a detector for the only category concerned. However, as a stand-alone face detector, RPN is not able to obtain competitive performances. We argue that such unsatisfactory performance comes from two aspects. Firstly, the anchors in the RPN are only associated with the last convolutional layer whose feature and resolution are too weak to handle faces of various sizes. Secondly, an anchor-associated layer is responsible for detecting faces within a corresponding range of scales, but it only has a single receptive field that can not match different scales of faces. To solve the above two problems, our MSCL is designed along the following two dimensions:

\setlength{\parskip}{-0.3\baselineskip}
\begin{itemize}
\setlength{\itemsep}{0pt}
\setlength{\parsep}{0pt}
\setlength{\parskip}{2pt}
\item \textbf{Multi-scale design along the dimension of network depth.} As shown in Fig.~\ref{fig:FaceBoxes},
our designed MSCL consists of several layers. These layers decrease in size progressively and form the multi-scale feature maps. Similar to~\cite{liu2016ssd}, our default anchors are associated with multi-scale feature maps (\textit{i.e.}, Inception3, Conv3\_2 and Conv4\_2). These layers, as a multi-scale design along the dimension of network depth, discretize anchors over multiple layers with different resolutions to naturally handle faces of various sizes. 
\item \textbf{Multi-scale design along the dimension of network width.} To learn visual patterns for different scales of faces, output features of the anchor-associated layers should correspond to various sizes of receptive fields, which can be easily fulfilled via Inception modules~\cite{szegedy2015going}. The Inception module consists of multiple convolution branches with different kernels. These branches, as a multi-scale design along the dimension of network width, is able to enrich the receptive fields. As shown in Fig.~\ref{fig:FaceBoxes}, the first three layers in MSCL are based on the Inception module.  Fig.~\ref{fig:Inception} illustrates our Inception implementation, which is a cost-effective module to capture different scales of faces.
\end{itemize}
\setlength{\parskip}{0\baselineskip} 

\subsection{Anchor densification strategy}
As illustrated in Fig.~\ref{fig:FaceBoxes}, we impose $1$:$1$ aspect ratio for the default anchors (i.e., square anchor), because the face box is approximately square. The scale of anchor for the Inception3 layer is $32$, $64$ and $128$ pixels, for the Conv3\_2 layer and Conv4\_2 layer are $256$ and $512$ pixels, respectively. 

The tiling interval of anchor on the image is equal to the stride size of the corresponding anchor-associated layer. For example, the stride size of Conv3\_2 is $64$ pixels and its anchor is $256\times256$, indicating that there is a $256\times256$ anchor for every $64$ pixels on the input image. We define the tiling density of anchor (\textit{i.e.}, $A_{density}$) as follows:

\vspace{-0.2cm}
\begin{equation}\label{equ:1}
A_{density} = A_{scale} / A_{interval}
\end{equation}
Here, $A_{scale}$ is the scale of anchor and $A_{interval}$ is the tiling interval of anchor. The tiling intervals for our default anchors are $32$, $32$, $32$, $64$ and $128$, respectively. According to Equ.~(\ref{equ:1}), the corresponding densities are $\bf{1}$, $\bf{2}$, $4$, $4$ and $4$, where it is obviously that there is a tiling density imbalance problem between anchors of different scales. Comparing with large anchors (\textit{i.e.}, $128\times128$, $256\times256$ and $512\times512$), small anchors (\textit{i.e.}, $32\times32$ and $64\times64$) are too sparse, which results in low recall rate of small faces. 

\setlength{\parskip}{0\baselineskip}
To eliminate this imbalance, we propose a new anchor densification strategy. Specifically, to densify one type of anchors $n$ times, we uniformly tile $A_{number}=n^2$ anchors around the center of one receptive field instead of only tiling one at the center of this receptive field to predict. Some examples are shown in Fig.~\ref{fig:AFW}. In our paper, to improve the tiling density of the small anchor, our strategy is used to densify the $32\times32$ anchor $4$ times and the $64\times64$ anchor $2$ times, which guarantees that different scales of anchor have the same density (\textit{i.e.}, 4) on the image, so that various scales of faces can match almost the same number of anchors.

\begin{figure}[htbp!]
\centering
\includegraphics[width=8cm]{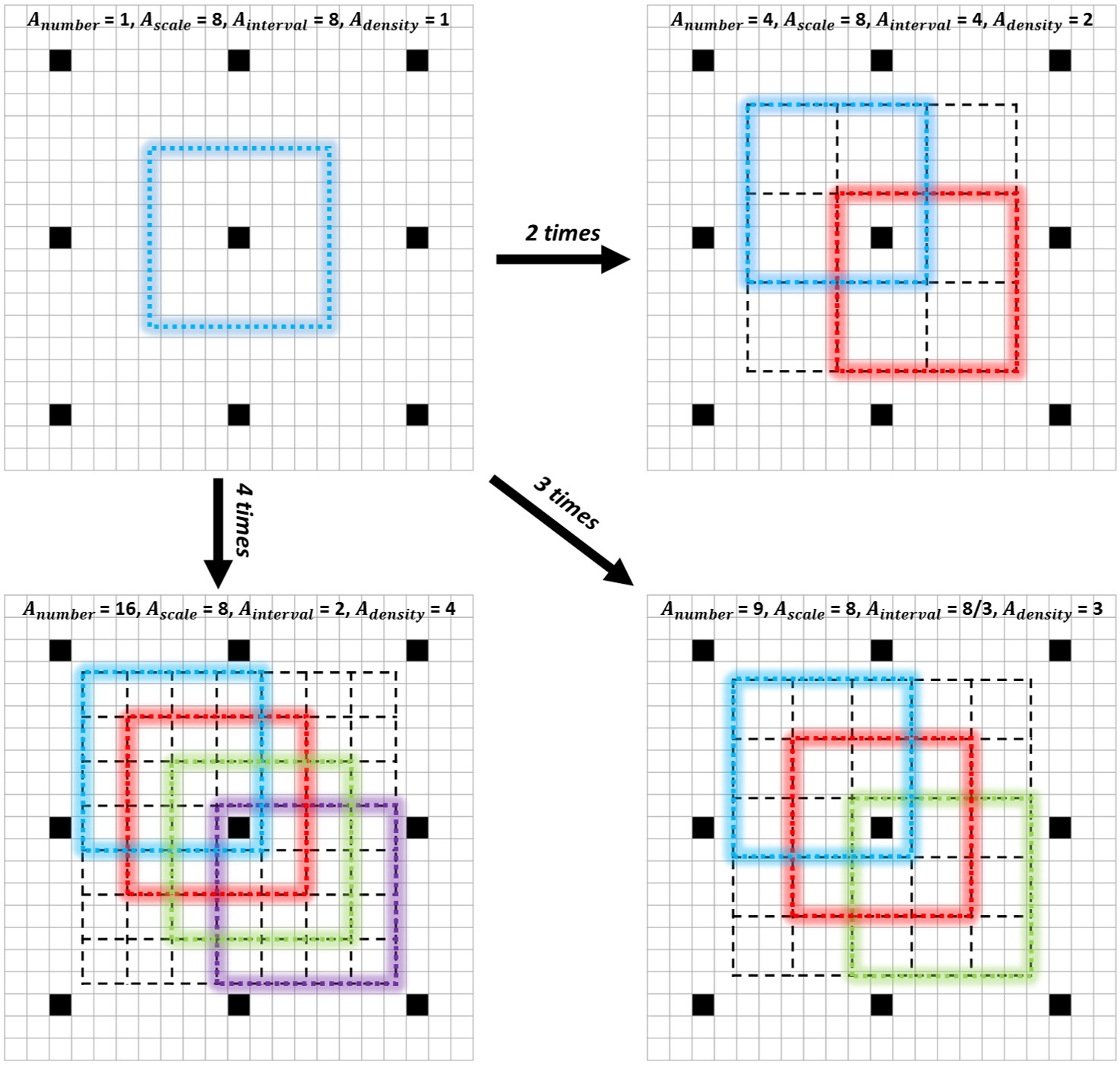}
\caption{Examples of anchor densification. For clarity, we only densify anchors at one receptive field centre (\textit{i.e.}, the central black cell), and only color the diagonal anchors.}
\label{fig:densification}
\end{figure}

\subsection{Training} \label{3.4}
This subsection introduces the training dataset, data augmentation, matching strategy, loss function, hard negative mining, and other implementation details.

{\bf Training dataset.} Our model is trained on $12,880$ images of the WIDER FACE~\cite{yang2016wider} training subset.

{\bf Data augmentation.} Each training image is sequentially processed by the following data augmentation strategies:
\setlength{\parskip}{-0.5\baselineskip}
\begin{itemize}
\setlength{\itemsep}{0pt}
\setlength{\parsep}{0pt}
\setlength{\parskip}{0pt}
\item Color distortion: Applying some photo-metric distortions similar to~\cite{howard2013some}.
\item Random cropping: We randomly crop five square patches from the original image: one is the biggest square patch, and the size of the others range between [$0.3$, $1$] of the short size of the original image. Then we arbitrarily select one patch for subsequent operations.
\item Scale transformation: After random cropping, the selected square patch is resized to $1024\times1024$.
\item Horizontal flipping: The resized image is horizontally flipped with probability of $0.5$.
\item Face-box filter: We keep the overlapped part of the face box if its center is in the above processed image, then filter out these face boxes whose height or width is less than 20 pixels.
\end{itemize}
\setlength{\parskip}{-0.5\baselineskip}

{\bf Matching strategy.} During training, we need to determine which anchors correspond to a face bounding box. We first match each face to the anchor with the best jaccard overlap, and then match anchors to any face with jaccard overlap higher than a threshold (\textit{i.e.}, $0.35$).
\setlength{\parskip}{0\baselineskip}

{\bf Loss function.} Our loss function is the same as RPN in Faster R-CNN~\cite{ren2015faster}. We adopt a 2-class softmax loss for classification and the smooth L1 loss for regression.

{\bf Hard negative mining.} After the anchor matching step, most of the anchors are found to be negative, which introduces a significant imbalance between the positive and negative examples. For faster optimization and stable training, we sort them by the loss values and pick the top ones so that the ratio between the negatives and positives is at most $3$:$1$.

{\bf Other implementation details.} All the parameters are randomly initialized with the ``xavier" method. We fine-tune the resulting model using SGD with $0.9$ momentum, $0.0005$ weight decay and batch size $32$. The maximum number of iterations is $120k$ and we use $10^{-3}$ learning rate for the first $80k$ iterations, then continue training for $20k$ iterations with $10^{-4}$ and $10^{-5}$, respectively. Our method is implemented in the Caffe library.

\section{Experiments} \label{4}
In this section, we firstly introduce the runtime efficiency of FaceBoxes, then analyze our model  in an ablative way, finally evaluate it on the common face detection benchmarks.

\subsection{Runtime efficiency}
CNN based methods have always been accused of its runtime efficiency. Although the existing CNN face detectors can be accelerated via high-end GPUs, they are not fast enough in most practical applications, especially CPU based applications. As described below, our FaceBoxes is efficient enough to meet practical requirements.

\setlength{\parskip}{-0.0\baselineskip}
During inference, our method outputs a large number of boxes (\textit{e.g.}, $8,525$ boxes for a VGA-resolution image). We first filter out most boxes by a confidence threshold of $0.05$ and keep the top $400$ boxes before applying NMS, then we perform NMS with jaccard overlap of $0.3$ and keep the top $200$ boxes. We measure the speed using Titan X (Pascal) and cuDNN v$5.1$ with Intel Xeon E$5$-$2660$v$3$@$2.60$GHz. As listed in Tab.~\ref{tab:time}, comparing with recent CNN-based methods, our FaceBoxes can run at $20$ FPS on the CPU with state-of-the-art accuracy. Besides, our method can run at $125$ FPS using a single GPU and has only $4.1$MB in size.

\vspace{-0.2cm}
\begin{table}[!htbp]
\setlength{\abovecaptionskip}{0.cm}
\setlength{\belowcaptionskip}{-0.5cm}
\centering
\begin{tabular}{cccc}
\toprule[2pt]
{\bf Approach} & {\bf CPU-model} & {\bf mAP(\%)} & {\bf FPS}\\
\midrule[1pt]
ACF~\cite{yang2014aggregate} & i7-3770@3.40 & 85.2 & 20\\
CasCNN~\cite{li2015convolutional} & E5-2620@2.00 & 85.7 & 14\\
FaceCraft~\cite{qin2016joint} & N/A & 90.8 & 10\\
STN~\cite{chen2016supervised} & i7-4770K@3.50 & 91.5 & 10\\
MTCNN~\cite{zhang2016joint} & N/A@2.60 & 94.4 & 16\\
Ours & E5-2660v3@2.60 & \textbf{96.0} & \textbf{20}\\
\bottomrule[2pt]
\end{tabular}
\vspace{2mm}
\caption{Overall CPU inference time and mAP compared on different methods. The \textbf{FPS} is for VGA-resolution images on CPU and the \textbf{mAP} means the true positive rate at $1000$ false positives on FDDB. Notably, for STN~\cite{chen2016supervised}, its mAP is the true positive rate at $179$ false positives and with ROI convolution, its FPS can be accelerated to $30$ with $0.6\%$ recall rate drop.}\label{tab:time}
\end{table}

\setlength{\parskip}{-1.0\baselineskip}
\subsection{Model analysis}
\setlength{\parskip}{-0.15\baselineskip}
We carried out extensive ablation experiments on the FDDB dataset to analyze our model. Comparing with AFW and PASCAL face, FDDB is much more difficult so that analyzing our model on FDDB is convincing. For all the experiments, we use the same settings, except for specified changes to the components.
\setlength{\parskip}{-0.0\baselineskip}

{\bf Ablative Setting.} To better understand FaceBoxes, we ablate each component one after another to examine how each proposed component affects the final performance. 1) Firstly, we ablate the anchor densification strategy. 2) Then, we replace MSCL with three convolutional layers, which all have $3\times3$ kernel size and whose output number is the same as the first three Inception modules of MSCL. Meantime, we only associate the anchors with the last convolutional layer. 3) Finally, we take the place of C.ReLU with ReLU in RDCL. The ablative results are listed in Tab.~\ref{tab:ablation} and some promising conclusions can be summed up as follows:

\vspace{-0.25cm}
\begin{table}[!htbp]
\centering
\begin{tabular}{c|cccc}
\toprule[2pt]
\multicolumn{1}{c|}{\bf Contribution}&\multicolumn{4}{c}{\bf FaceBoxes}\\
\midrule[1pt]
RDCL & & & & {\bf \texttimes}\\
MSCL & & & {\bf \texttimes} & {\bf \texttimes}\\
Strategy & & {\bf \texttimes} & {\bf \texttimes} & {\bf \texttimes}\\
\midrule[1pt]
Accuracy (mAP) &96.0 & 94.9 & 93.9 & 94.0\\
Speed (ms) & 50.98 & 48.27 & 48.23 & 67.48 \\
\bottomrule[2pt]
\end{tabular}
\hspace{1in}
\caption{Ablative results of the FaceBoxes on FDDB dataset. Accuracy (mAP) means the true positive rate at $1000$ false positives. Speed (ms) is for the VGA-resolution images on the CPU.}\label{tab:ablation}
\end{table}

\setlength{\parskip}{-0.8\baselineskip}
{\bf Anchor densification strategy is crucial.} Our anchor densification strategy is used to increase the density of small anchors (\textit{i.e.}, $32\times32$ and $64\times64$) in order to improve the recall rate of small faces. From the results listed in Tab.~\ref{tab:ablation}, we can see that the mAP on FDDB is reduced from $96.0\%$ to $94.9\%$ after ablating the anchor densification strategy. The sharp decline (\textit{i.e.}, $1.1\%$) demonstrates the effectiveness of the proposed anchor densification strategy.

\setlength{\parskip}{-0.0\baselineskip}
{\bf MSCL is better.} The comparison between the second and third columns in Tab.~\ref{tab:ablation} indicates that MSCL effectively increases the mAP by $1.0\%$, owning to the diverse receptive fields and the multi-scale anchor tiling mechanism.

{\bf RDCL is efficient and accuracy-preserving.} The design of RDCL enables our FaceBoxes to achieve real-time speed on the CPU. As reported in Tab.~\ref{tab:ablation}, RDCL leads to a negligible decline on accuracy but a significant improvement on speed. Specifically, the FDDB mAP decreases by $0.1\%$ in return for the about $19.3$ms speed improvement. 
\subsection{Evaluation on benchmark}
\setlength{\parskip}{-0.25\baselineskip}
We evaluate the FaceBoxes on the common face detection benchmark datasets, including Annotated Faces in the Wild (AFW), PASCAL Face, and Face Detection Data Set and Benchmark (FDDB).

\begin{figure}[htbp]
\centering
\includegraphics[width=0.4\textwidth]{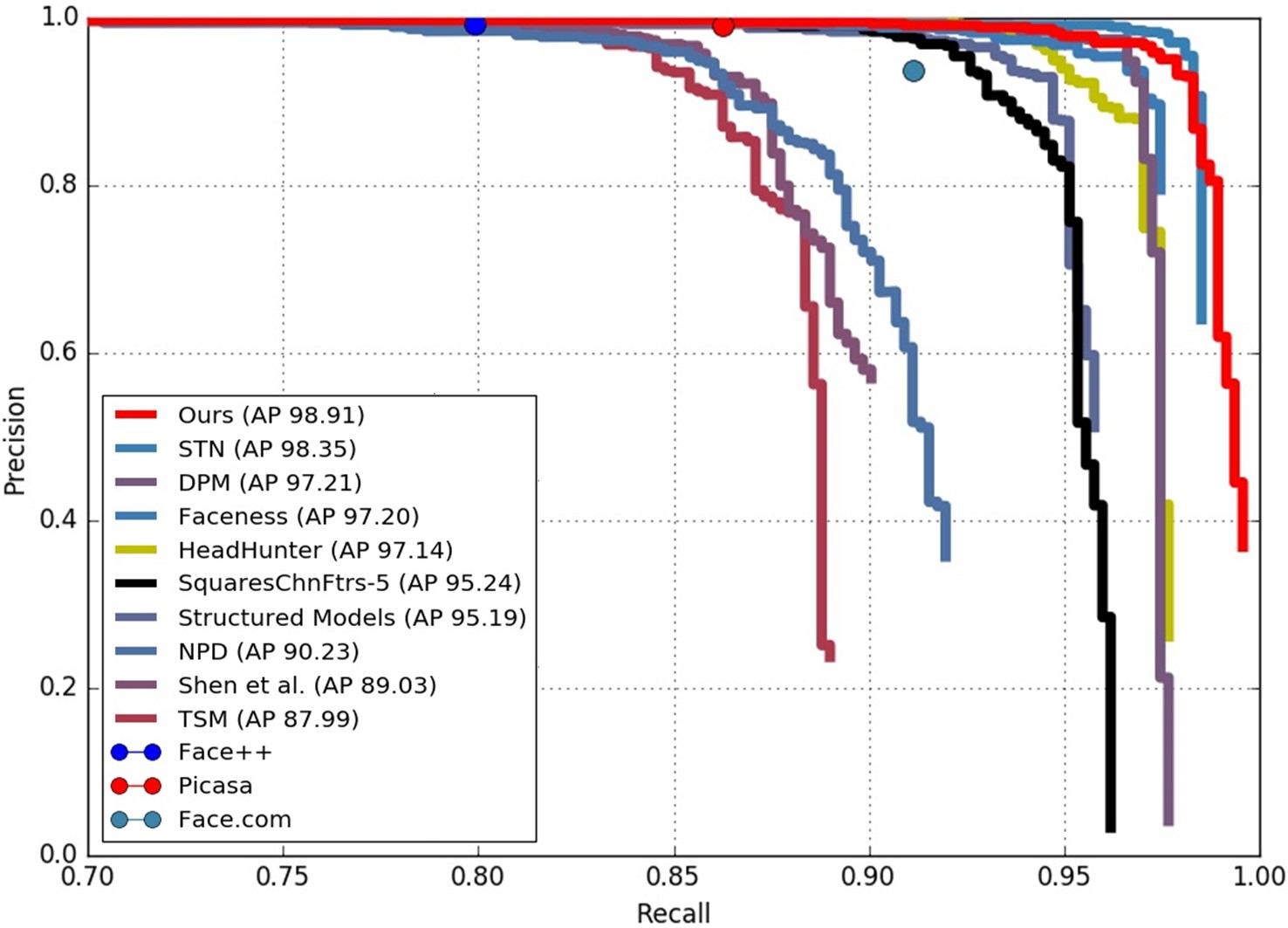}
\caption{Precision-recall curves on AFW dataset.}
\label{fig:AFW}
\end{figure}

\begin{figure*}[htbp]
\centering
\subfigure[AFW]{
\label{fig:afw-q} 
\includegraphics[width=0.925\textwidth]{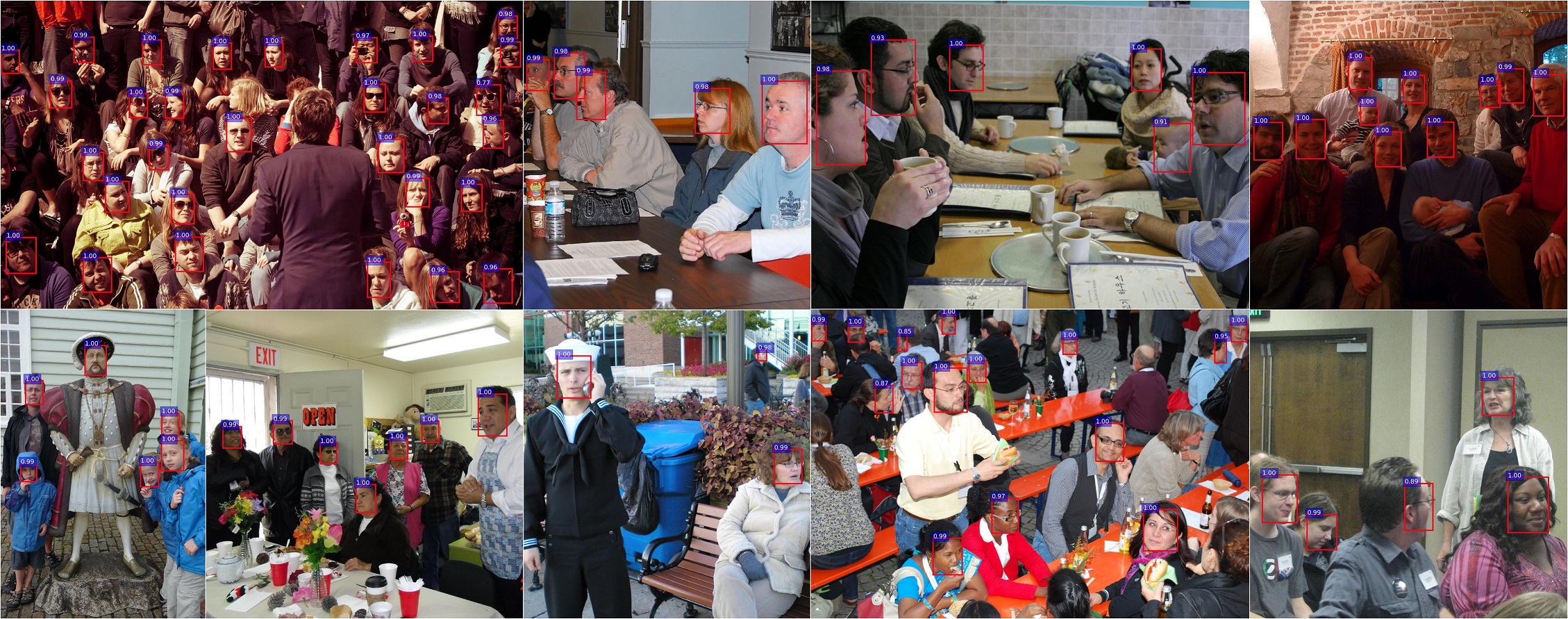}}
\subfigure[PASCAL face]{
\label{fig:pascal-q} 
\includegraphics[width=0.925\textwidth]{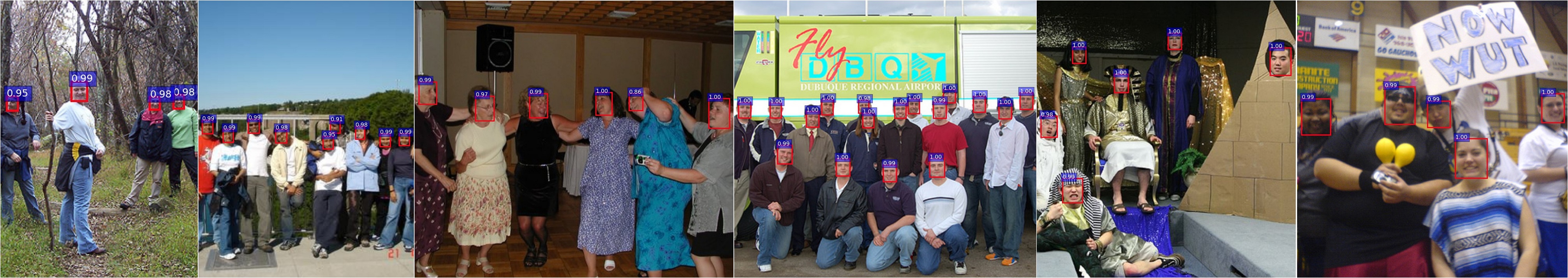}}
\subfigure[FDDB]{
\label{fig:fddb-q} 
\includegraphics[width=0.925\textwidth]{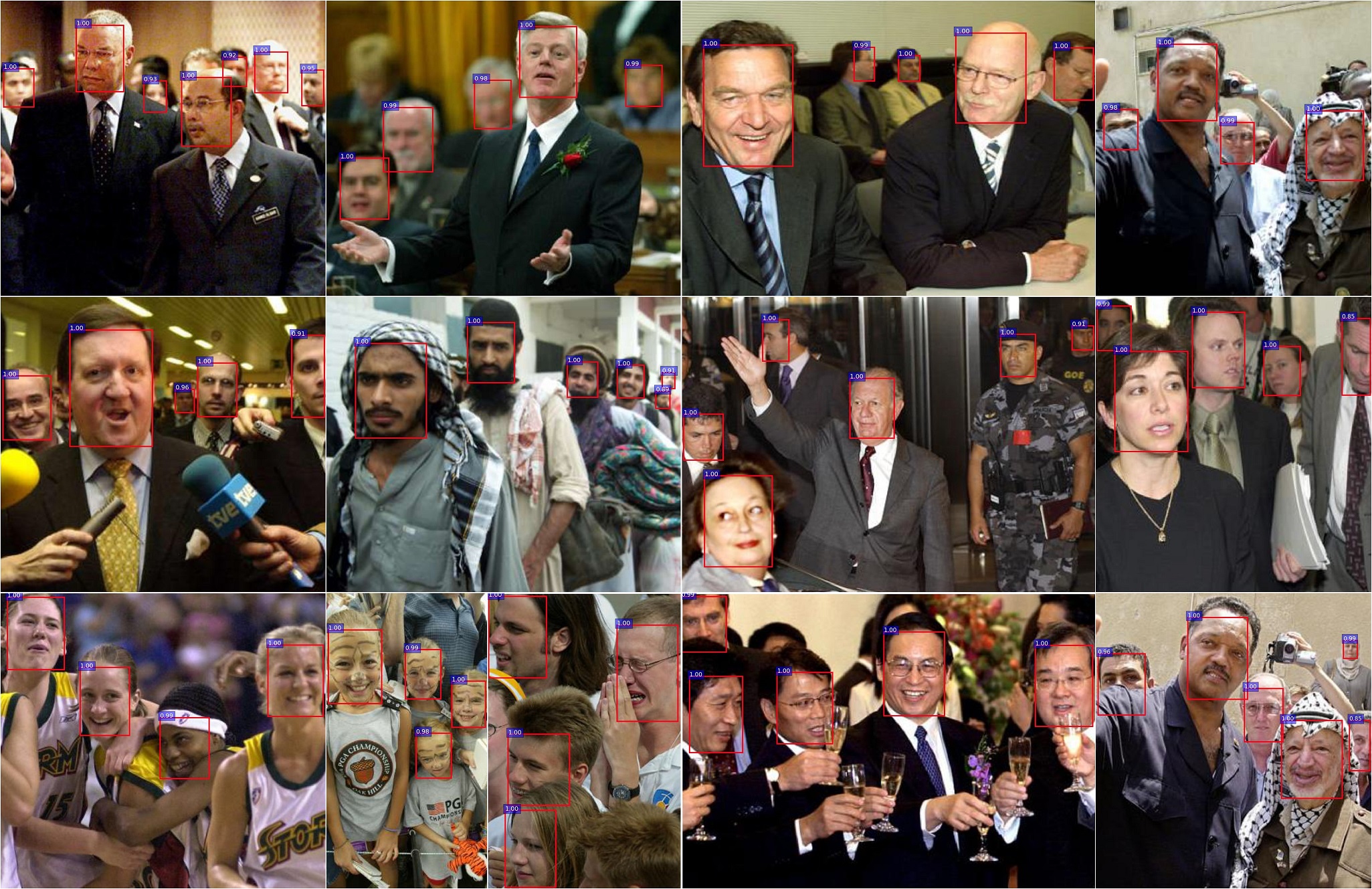}}
\caption{Qualitative results on face detection benchmark datasets.}
\label{fig:q-r} 
\end{figure*}

\setlength{\parskip}{-0.0\baselineskip}
{\bf AFW dataset}~\cite{zhu2012face}. It has $205$ images with $473$ faces. We evaluate FaceBoxes against the well-known works~\cite{chen2016supervised,liao2016fast,mathias2014face,shen2013detecting,yan2014face,yang2015facial,zhu2012face} and commercial face detectors (\textit{e.g.}, Face.com, Face++ and Picasa). As illustrated in Fig.\ref{fig:AFW}, our FaceBoxes outperforms all others by a large margin. Fig.\ref{fig:afw-q} shows some qualitative results on the AFW dataset.

\begin{figure}[htbp]
\centering
\includegraphics[width=0.4\textwidth]{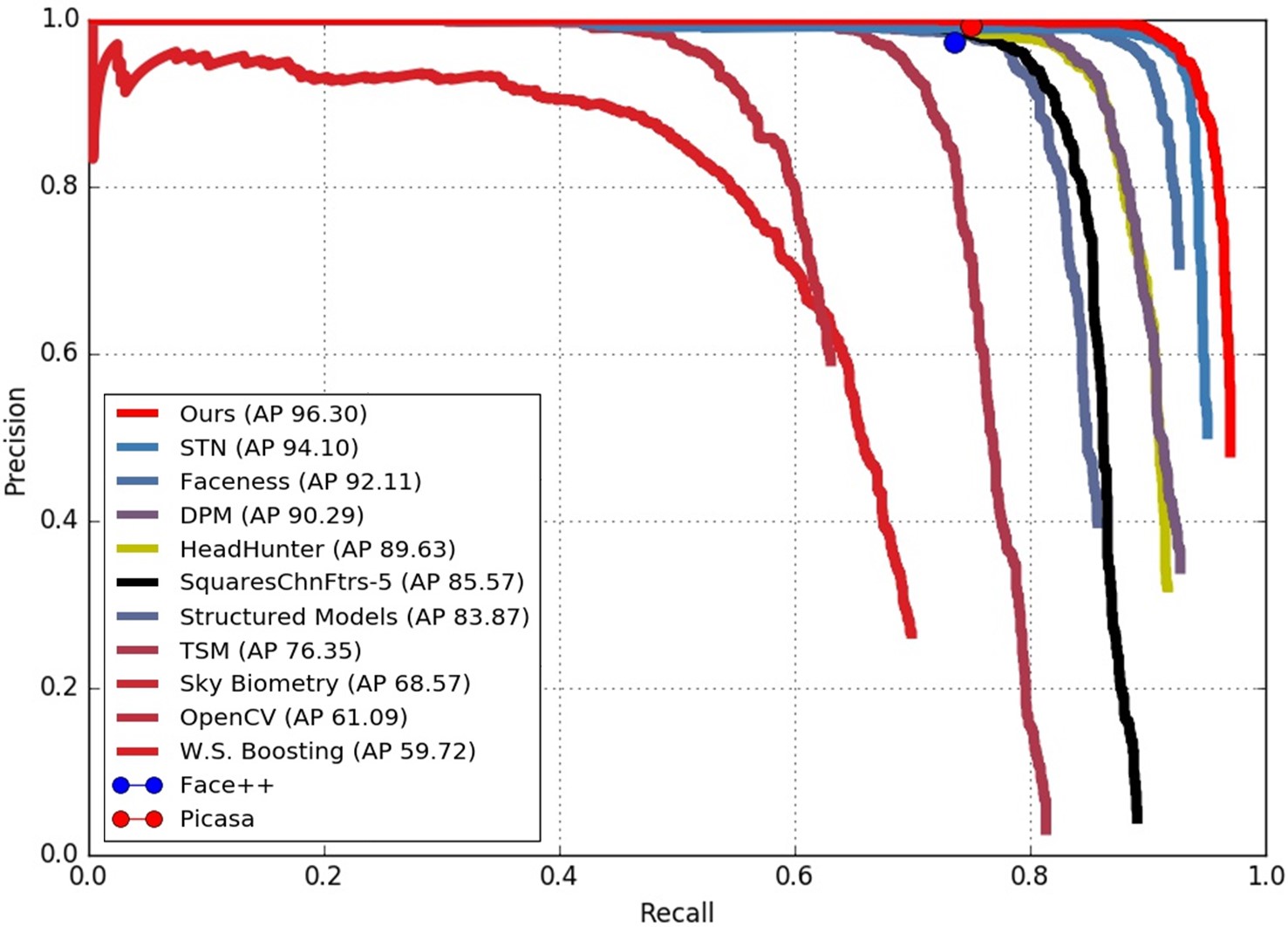}
\caption{Precision-recall curves on PASCAL face dataset.}
\label{fig:PASCAL}
\end{figure}

\setlength{\parskip}{-0.65\baselineskip}
{\bf PASCAL face dataset}~\cite{yan2014face}. It is collected from the test set of PASCAL person layout dataset, consisting of $1335$ faces with large face appearance and pose variations from $851$ images. Fig.\ref{fig:PASCAL} shows the precision-recall curves on this dataset. Our method significantly outperforms all other methods~\cite{chen2016supervised,kalal2008weighted,mathias2014face,yan2014face,yang2015facial,zhu2012face} and commercial face detectors (\textit{e.g.}, SkyBiometry, Face++ and Picasa). Fig.\ref{fig:pascal-q} shows some qualitative results on the PASCAL face dataset.

\begin{figure}[htbp]
\centering
\subfigure[Discontinuous ROC curves]{
\label{fig:FDDBd}
\includegraphics[width=0.425\textwidth]{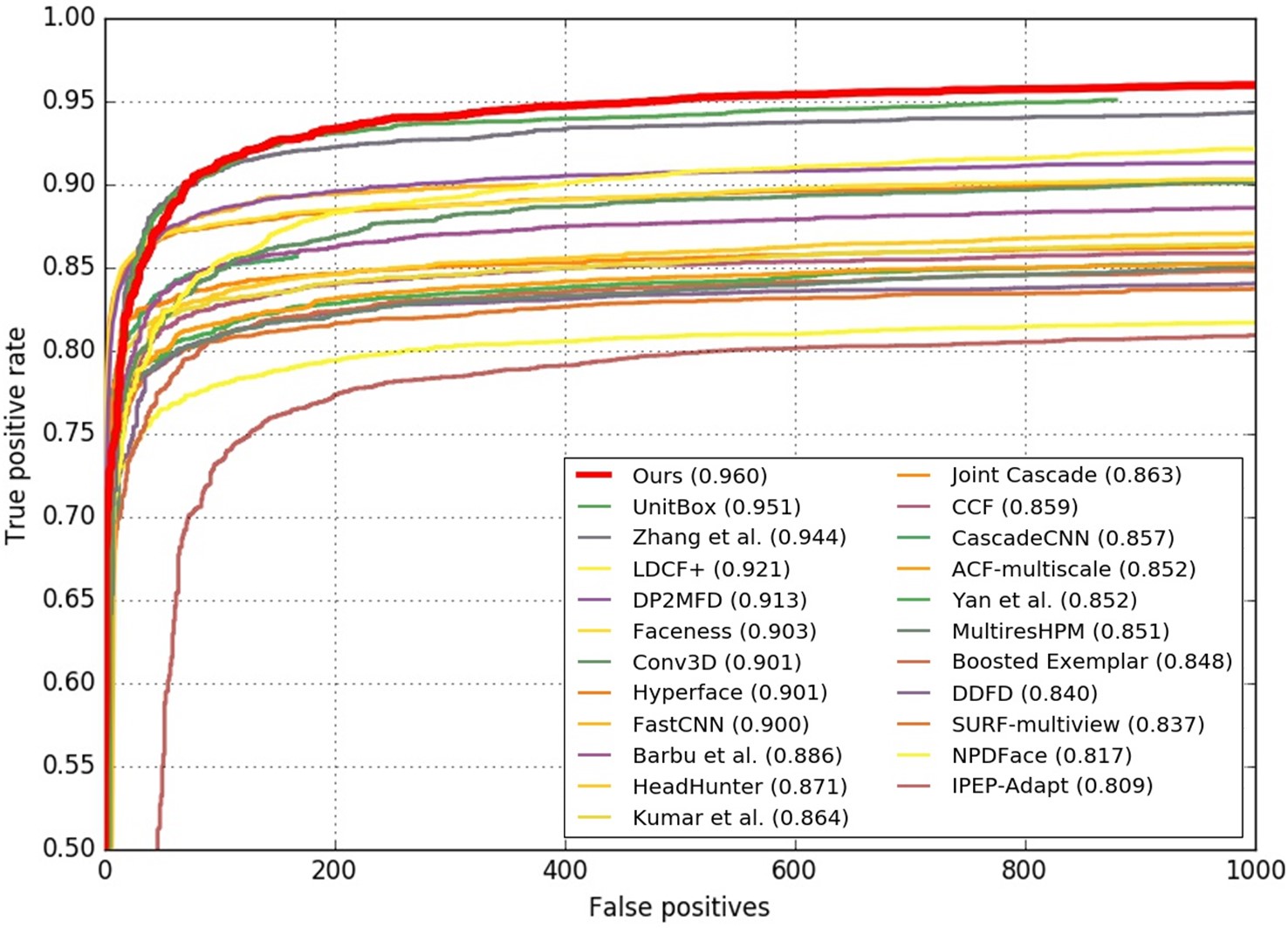}}
\subfigure[Continuous ROC curves]{
\label{fig:FDDBc} 
\includegraphics[width=0.425\textwidth]{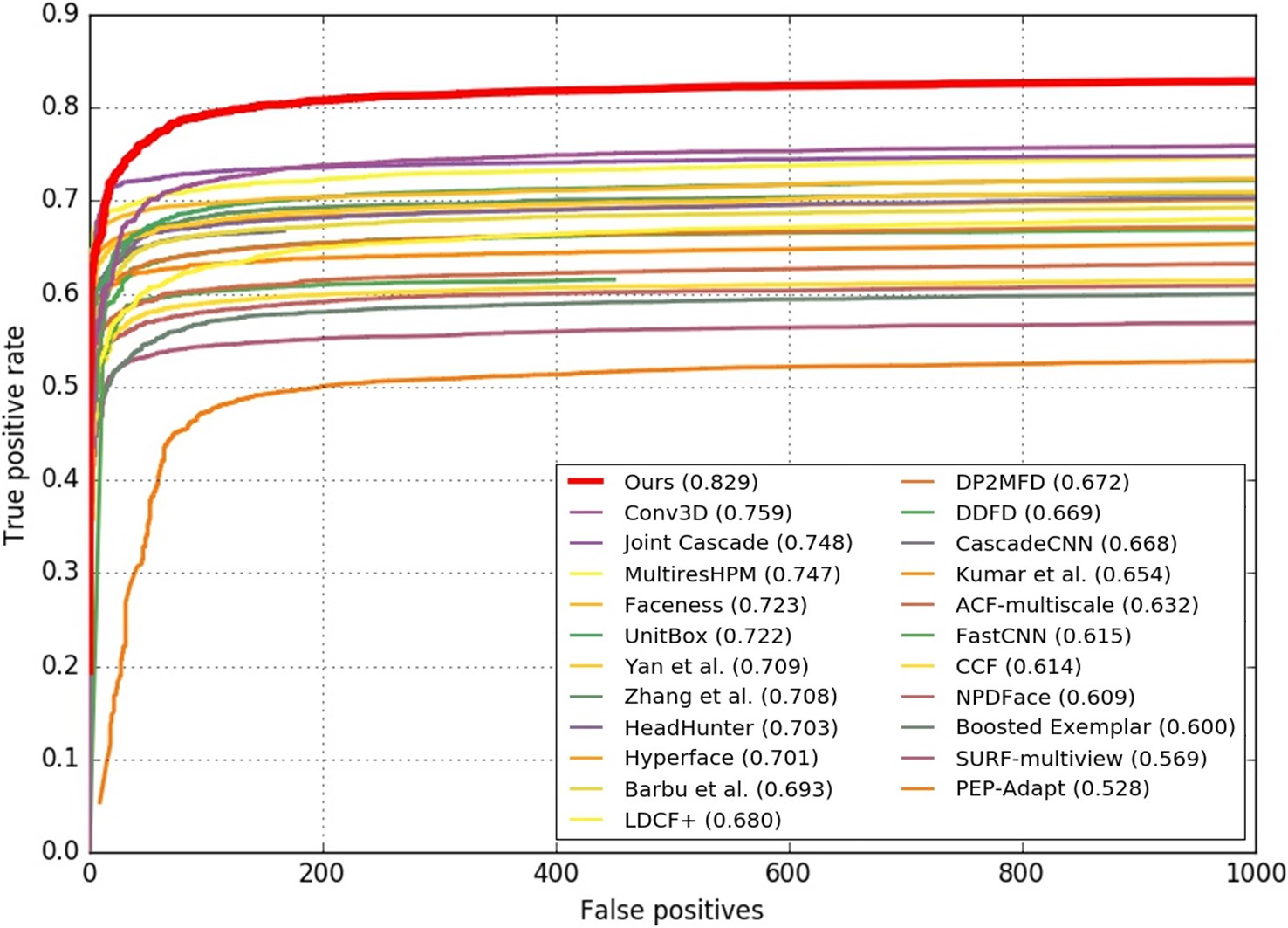}}
\caption{Evaluation on the FDDB dataset.}
\label{fig:Evaluation}
\end{figure}

\setlength{\parskip}{-1.2\baselineskip}
{\bf FDDB dataset}~\cite{jain2010fddb}. It has $5,171$ faces in $2,845$ images taken from news articles on Yahoo websites. FDDB adopts the bounding ellipse, while our FaceBoxes outputs rectangle bounding box. This inconsistency has a great impact to the continuous score. For a more fair comparison under the continuous score evaluation, we train an elliptical regressor to transform our predicted bounding boxes to bounding ellipses. We evaluate our face detector on FDDB against the other methods~\cite{barbu2014face,farfade2015multi,ghiasi2015occlusion,kumar2015visual,li2013probabilistic,li2014efficient,li2013learning,li2016face,liao2016fast,ohn2017boost,ranjan2015deep,ranjan2016hyperface,triantafyllidou2016fast,yang2015facial,yu2016unitbox,zhang2016joint}. The results are shown in Fig.~\ref{fig:FDDBd} and Fig.\ref{fig:FDDBc}. Our FaceBoxes achieves the state-of-the-art performance and outperforms all others by a large margin on discontinuous and continuous ROC curves. These results indicate that our FaceBoxes can robustly detect unconstrained faces. Fig.\ref{fig:fddb-q} shows some qualitative results on the FDDB.
\setlength{\parskip}{-0.0\baselineskip}

\setlength{\parskip}{-0.75\baselineskip}
\section{Conclusion} \label{5}
\setlength{\parskip}{-0.4\baselineskip}
Since effective models for the face detection task tend to be computationally prohibitive, it is challenging for the CPU devices to achieve real-time speed as well as maintain high performance. In this work, we present a novel face detector with superior performance on both speed and accuracy. The proposed method has a lightweight yet powerful network structure, which consists of RDCL and MSCL. The former enables FaceBoxes to achieve real-time speed, and the latter aims at enriching receptive fields and discretizing anchors over different layers to handle faces of various scales. Besides, a new anchor densification strategy is proposed to improve the recall rate of small faces. The experiments demonstrate that our contributions lead FaceBoxes to the state-of-the-art performance on the common face detection benchmarks. The proposed detector is very fast, achieving $20$ FPS for VGA-resolution images on CPU and can be accelerated to $125$ FPS on GPU.
\setlength{\parskip}{-0.0\baselineskip}

\section*{Acknowledgments} \label{6}
This work was supported by the National Key Research and Development Plan (Grant No.2016YFC0801002), the Chinese National Natural Science Foundation Projects $\#61473291$, $\#61502491$, $\#61572501$, $\#61572536$, $\#61672521$ and AuthenMetric R\&D Funds.

\begin{spacing}{1.0}
{\small
\bibliographystyle{ieee}
\bibliography{submission_example}
}
\end{spacing}

\end{document}